\documentclass{article}
\usepackage{spconf,amsmath,graphicx}
\usepackage{multirow}
\usepackage{booktabs}
\usepackage{url}
\usepackage{graphicx}
\newcommand{\mycomment}[1]{}
\usepackage{booktabs}
\usepackage{algorithm}
\usepackage{algorithmicx}   
\usepackage{algpseudocode} 
\usepackage{xcolor}
\newcommand{\predimg}[1]{%
  \includegraphics[width=1.2cm,height=1.2cm,keepaspectratio]{#1}}
\usepackage{enumitem}
\usepackage{setspace}
\usepackage{stfloats}
\usepackage{placeins}
\title{M\lowercase{u}CALD-SplitFed: Causal-Latent Diffusion for Privacy-Preserving Multi-Task Split-Federated Medical Image Segmentation}
\name{Chamani Shiranthika, Hadi Hadizadeh, and Parvaneh Saeedi
\thanks{Thanks to Natural Sciences and Engineering Research Council (NSERC) of Canada for funding.}
}
\address{School of Engineering Science, Simon Fraser University, Burnaby, BC, Canada}
\begin{document}
%\ninept
%
\maketitle
\begin{abstract}
Federated Learning enables decentralized training by aggregating model updates across clients without sharing raw data, while Split Federated Learning further partitions the model between clients and a server to reduce computation and communication at the client side. However, decentralized medical institutions rarely operate on a single shared task, making standard Federated and SplitFed collaborations poorly aligned with real clinical workflows. Multi-task FL extends these frameworks by allowing clients to handle different tasks, but often introduces instability and privacy vulnerabilities. This study proposes \textbf{MuCALD-SplitFed}, a multi-task SplitFed framework that integrates causal representation learning and latent diffusion. Experiments show MuCALD-SplitFed consistently improves segmentation, while baseline SplitFed fails to converge. The proposed approach further reduces information leakage at split points, mitigating reconstruction-based and membership inference attacks. Additionally, MuCALD SplitFed outperforms state-of-the-art personalized FL and multi-task FL approaches. The code repository is: \url{https://github.com/ChamaniS/MuCALD_SplitFed}.

\end{abstract}
\section{Introduction}
\label{sec:intro}
Medical image segmentation plays a critical role in clinical workflows. Hospitals, clinics, and research laboratories often differ in imaging modalities, annotation quality, acquisition protocols, population demographics, etc. Practical medical AI systems therefore require learning frameworks that can generalize across heterogeneous tasks while preserving privacy. Decentralized learning paradigms such as Federated Learning (FL) \cite{mcmahan_2017}, Split Learning (SL) \cite{Gupta_2018}, and Split Federated Learning (SplitFed) \cite{thapa2022splitfed} enable collaborative model training without sharing raw data. However, these approaches implicitly assume all clients solve the same task—the non-IID assumption \cite{mcmahan_2017}. which is unrealistic in real clinical settings. %where institutions perform distinct tasks. 

Multi-task learning relaxes this constraint by allowing clients to train on domain-specific tasks while benefiting from shared global representations. In the literature, \textit{multi-task} refer to clients with non-IID institutional data, or clients working on related but different tasks (which is neither IID nor non-IID), or setups where multiple objectives (e.g., segmentation and classification) are learned jointly. Multi-task FL methods- MOCHA \cite{smith2017federated}, sheaf-based FL \cite{issaid2025tackling}, FedAlign \cite{gupta2025fedalign}, and multi-task SplitFed variants- FedBone \cite{chen2024fedbone}, FedMSplit \cite{chen2022fedmsplit}, homomorphically encrypted multi-task SL \cite{dong2025multi} demonstrate effective collaboration under heterogeneous tasks.

Multi-task SplitFed systems often introduce convergence instability, as clients optimize heterogeneous objectives across different data distributions and label spaces. During federated aggregation, gradients from different tasks can conflict, leading to unstable or oscillatory updates. The shared latent representation at split points entangles task-specific features, amplifying cross-task interference and hindering convergence. Multi-task SplitFed further introduces generalization and privacy challenges. Split-point communications can leak sensitive information, enabling attacks such as membership inference, reconstruction, property inference, and task-driven client drift. These risks are amplified by diverse task semantics in shared representations.

Addressing these vulnerabilities requires going beyond correlation-based representations. Causality explicitly models cause–effect relationships rather than statistical associations, enabling models to distinguish task-relevant structure from spurious correlations \cite{pearl2018book}. Recent research in causality in medical AI shows that causal representations improve robustness, reduce bias, and support trustworthy clinical decision-making \cite{castro2020causality}. Causality-driven models- C-CAM \cite{chen2022c}, CauSSL\cite{miao2023caussl}, and MACAW \cite{vigneshwaran2024macaw} demonstrate that modelling cause-effect structure yields stronger generalization under distribution shifts. Causal diffusion models— CausalDiffAE \cite{komanduri2024causal}, Diff-structural causal model (SCM) \cite{sanchez2022diffusion}, and latent-causal DDPMs \cite{zhangcausaldiff} indicate that combining causality and diffusion disentangles semantic factors, suppresses spurious correlations, and enables fine-grained control. However, these ideas have not been explored in multi-task SplitFed.

To address these gaps, we propose MuCALD-SplitFed, a causal, diffusion-driven multi-task SplitFed framework for privacy-preserving medical image segmentation. The contributions are: integrating causal modelling within the SplitFed latent space for multi-task segmentation in heterogeneous clinical settings; combining causal representations with latent diffusion and adversarial denoising to improve robustness and reduce information leakage at split points; disentangling task-relevant causal factors from domain-specific information to enhance segmentation performance and privacy; and extensive evaluation on five heterogeneous medical datasets %demonstrating consistent improvements 
over Baseline SplitFed and state-of-the-art multi-task and personalized FL methods. Section~\ref{sec:methodology} introduces the MuCALD-SplitFed methodology, Section~\ref{sec:experiments} presents experiments and SoTA comparisons, and Section~\ref{sec:conclusion} concludes the paper. 

\section{Proposed MuCALD-SplitFed}
\label{sec:methodology}
\begin{figure}
\centerline{\includegraphics[scale = 0.31]{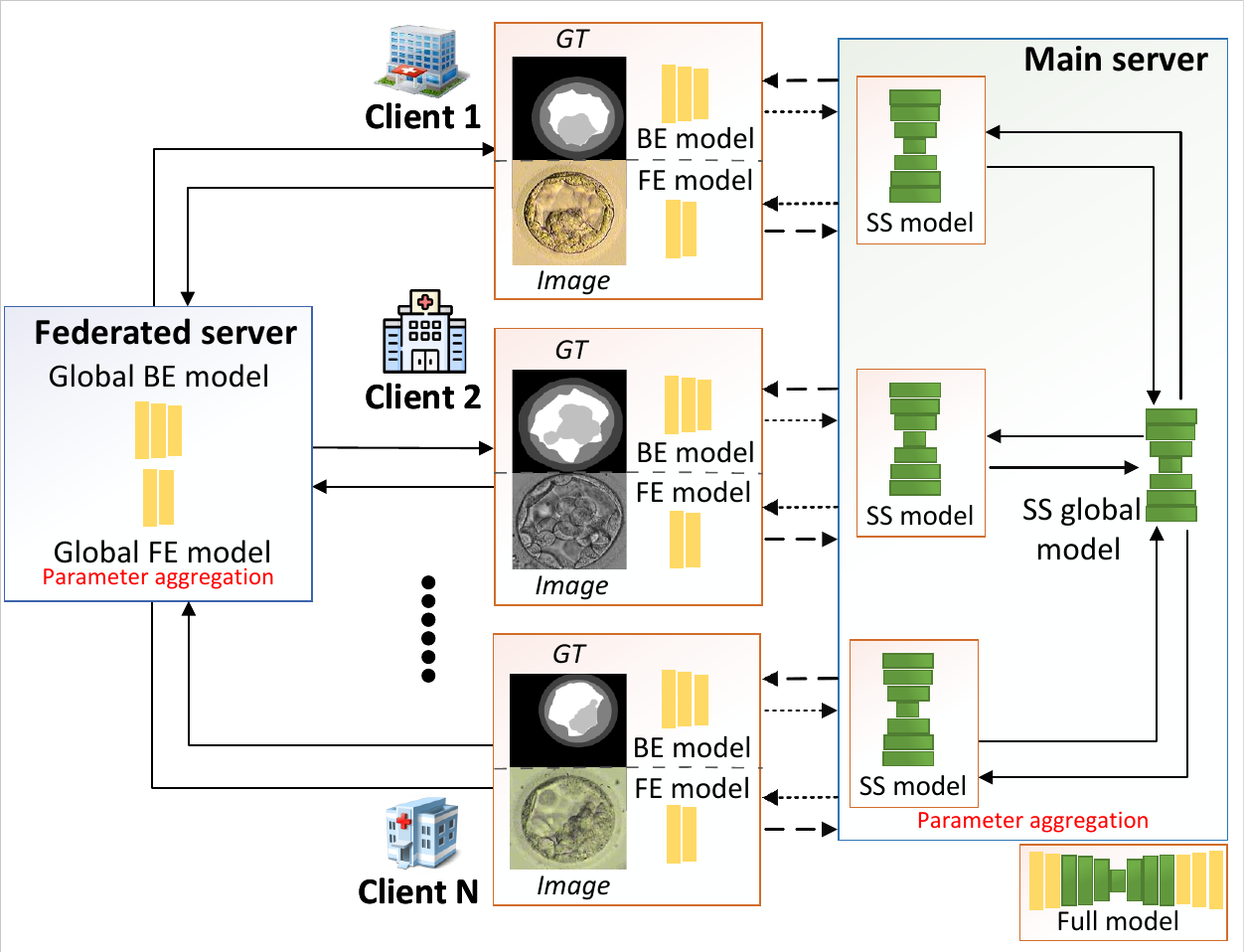}}
\caption{\small {Baseline SplitFed where all clients collaborate on a single shared task (blastocysts segmentation)}.}
\label{fig:baselinesplitfed}
\end{figure}

\begin{figure*}
\centerline{\includegraphics[scale = 0.61]{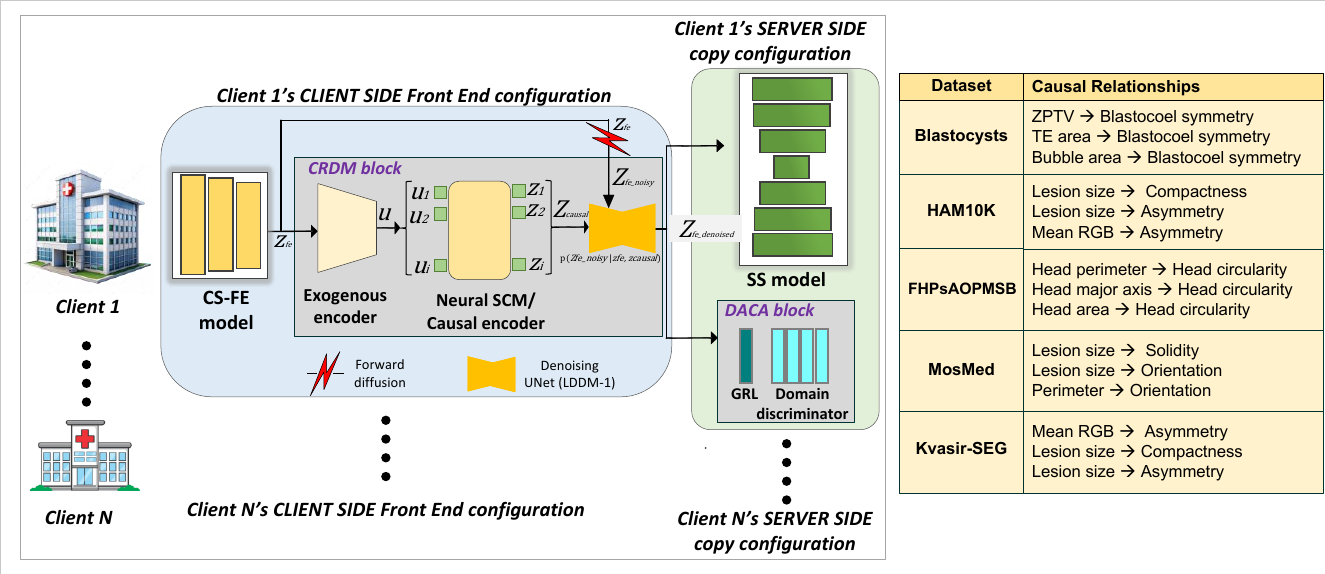}}
\caption{\small {Left: Proposed MuCALD SplitFed (CS-FE to SS split point). Right: Causal relationships discovered using Notears-MLP \cite{zheng2020learning} for Neural SCM modeling. A symmetric configuration operates at the SS to CS-BE split point.}}
\label{fig:mucaldsplitfed}
\end{figure*}

Our Baseline SplitFed architecture,~\cite{shiranthika_2023, Shiranthika_AASFL_2024},
based on~\cite{thapa2022splitfed} (Fig.~\ref{fig:baselinesplitfed}), partitions a model into front-end (FE), server-side (SS), and back-end (BE) components. Clients receive copies of the global FE and BE models from the federated server and the SS model from the main server. Each client then trains locally for several epochs, after which parameters are aggregated via federated averaging~\cite{mcmahan_2017} and redistributed. This  process repeats for multiple rounds until convergence. Proposed MuCALD-SplitFed framework (left side of Fig.~\ref{fig:mucaldsplitfed}) consists of three stages: (1) proxy-label construction and causal graph discovery, (2) causal representation and diffusion module (CRDM), and (3) domain-adversarial causal alignment (DACA). Each stage improves robustness, reduces leakage, or mitigates domain shift, respectively, without added cross-client dependencies.

\subsection{Proxy-label construction and causal graph discovery}
First, domain-specific morphological and intensity features are extracted from the five datasets as proxy labels for weakly supervised causal learning.
\begin{itemize}
    \item Blastocyst \cite{lockhart_2019}: zona pellucida thickness variation (ZPTV), ZP area, blastocoel (BL) area, blastocyst diameter, BL symmetry, inner cell mass (ICM) area/ solidity/ compactness/ entropy/ brightness, trophectoderm (TE) area/brightness/granularity, embryo expansion grade, bubble count/area, and ZP sharpness.
    \item  FHPsAOPMSB \cite{lu2022jnu}: fetal head size, perimeter, major/ minor axes, circularity, solidity, intensity statistics, bounding-box dimensions, symphysis features, head–symphysis geometric relations, and vertical alignment.
    \item HAM10K \cite{tschandl_2018}, MosMed \cite{morozov2020mosmeddata}, and Kvasir-SEG \cite{jha2020_kvasir}: lesion size, perimeter, compactness, bounding-box size, orientation, solidity, asymmetry, intensity statistics, RGB/ HSV statistics, colour variance, and entropy.
\end{itemize}

Dataset-specific causal graphs are then learned from these proxy features using Notears-MLP~\cite{zheng2020learning}. Three strongest causal relations per dataset are selected (right side of Fig.~\ref{fig:mucaldsplitfed}).

\subsection{Causal Representation and Diffusion Module (CRDM)}
The CRDM block consists of the exogenous encoder, neural SCM, and the denoising LDDM, followed by a forward diffusion step. First the CS-FE model produces feature maps $z_{fe}$, which the exogenous encoder maps to low-dimensional exogenous variables $u=z_{exo}=f_{exo}(z_{fe}) $. Second, the Neural-SCM learns a causal factorization that produces causally structured latents ($z_{causal}=f_{SCM}{u}$). Third, forward diffusion is applied to the encoder features ($z_{fe\_noisy} = forward\_diffusion(z_{fe}) $), to obfuscate transmitted activations. Next, a causal diffusion decoder (LDDM-1) denoises the noisy latents $z_{fe\_noisy}$ conditioned on the causal latents $z_{causal}$, generating $z_{denoised}$, which will be sent to the server-side. These causally structured, diffusion-noised latents expose less reconstructive information than standard SplitFed activations, reducing reconstruction and membership inference risks.

\subsection{Domain-Adversarial Causal Alignment (DACA)}
The server-side DACA block enforces domain invariance and suppresses task-specific biases in the received latents. It comprises a gradient reversal layer (GRL) \cite{ganin2015unsupervised}, which inverts gradients flowing into the client, and a lightweight domain discriminator, which attempts to predict the originating client or domain. During training, the segmentation network attempts to fool the discriminator, producing domain-agnostic representations. The discriminator penalizes domain-specific leakage, improving privacy and cross-task generalization.

\subsection{Optimization objective}
The Neural-SCM is weakly supervised using proxy labels, yielding proxy-alignment losses $L_{\text{proxy}}^{(1)}$ and $L_{\text{proxy}}^{(2)}$ at the two split points. Local model training jointly minimizes proxy-alignment losses, segmentation loss $L_{\text{seg}}$, %($L_{\text{proxy}}^{(1)}$, $L_{\text{proxy}}^{(2)}$), 
diffusion regularization losses ($L_{\text{diff}}^{(1)}$ and $L_{\text{diff}}^{(2)}$), KL divergence terms for the exogenous and causal latents 
$(L_{\text{KL}u},\, L_{\text{KL}z})$, 
and the adversarial domain losses 
($L_{\text{adv}}^{(1)}$ and $L_{\text{adv}}^{(2)}$) as in Equation \ref{eq:compact_loss}.
\begin{equation}
\begin{aligned}
L \;=\;& 
\lambda_{\text{seg}}\, L_{\text{seg}}
+\lambda_{\text{proxy}} \left( L_{\text{proxy}}^{(1)} + L_{\text{proxy}}^{(2)} \right)
+\lambda_{\text{diff}} \left( L_{\text{diff}}^{(1)} + L_{\text{diff}}^{(2)} \right) \\
&\quad
+\lambda_{\text{KL}u}\, L_{\text{KL}u}
+\lambda_{\text{KL}z}\, L_{\text{KL}z}
+\lambda_{\text{adv}} \left( L_{\text{adv}}^{(1)} + L_{\text{adv}}^{(2)} \right).
\end{aligned}
\label{eq:compact_loss}
\end{equation}

After each communication round, the client copies of SS models, LDDMs, exogenous encoders, and Neural-SCMs are aggregated. FE and BE remain local, preserving personalization and preventing leakage of task-specific attributes.
\section{Experimental Results}
\label{sec:experiments}
\subsection{Datasets and model training}
Our multi-task SplitFed framework consists of 5 clients, each occupying a  medical imaging dataset. Client 1 uses \textbf{Blastocyst} dataset \cite{lockhart_2019} (781 RGB human day-5 embryo images annotated into ZP, TE, BL, ICM, and background). Client 2 holds \textbf{HAM10K} dataset \cite{tschandl_2018} (10,015 dermatoscopic RGB images for skin lesion and background segmentation). Client 3 uses \textbf{FHPsAOPMSB} dataset \cite{lu2022jnu} (4,000 intrapartum transperineal ultrasound images segmented into fetal head, pubic symphysis, and background). Client 4 uses \textbf{MosMed} dataset \cite{morozov2020mosmeddata} (2800 lung computed tomography scan images, segmented into the affected area and background). Client 5 holds \textbf{Kvasir-SEG} dataset \cite{jha2020_kvasir} (1,000 endoscopic RGB polyp images segmented into abnormal conditions (lesion/polyp/ulcer) and background). Each client has fixed test sets (70, 1,000, 800, 547, 100), with the remaining data split 85\%,/15\% for training/validation. The datasets span diverse modalities with significant domain shifts, while causal modeling reduces data dependency, supporting generalization. Augmentations include flips, rotations, RGB shifts, normalization, and brightness/contrast adjustments. Models were trained with soft dice loss~\cite{sudre_2017} and a cosine diffusion schedule. Evaluation used class wise-average intersection over union~\cite{Cox_2008} with and without background (IoU W/B \& IoU N/B), precision, recall, F1 score, hausdorff distance (HD95), and average symmetric surface distance (ASSD). We used 24 communication rounds with 5 local epochs per client; 2 warm-up epochs (segmentation only), 3 ramp-up epochs with increasing proxy-label weights, and a final epoch.
\subsection{Performance comparison}
\begin{figure}
\centerline{\includegraphics[scale = 0.19]{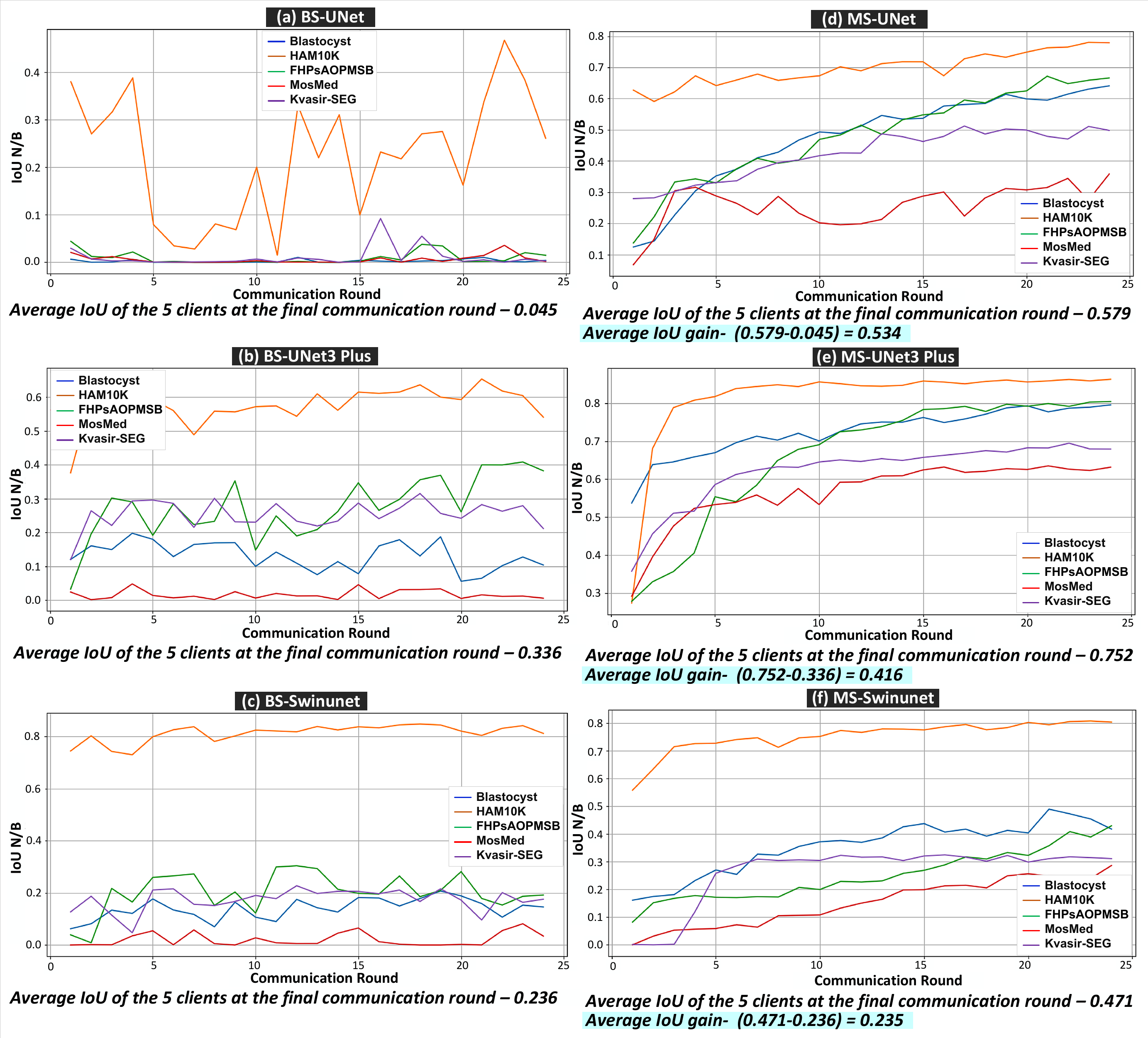}}
\caption{\small{IoU N/B trajectories across communication rounds for Baseline SplitFed (BS; a–c) and MuCALD-SplitFed (MS; d–f). The average IoU N/B at the final communication round is reported for each model, together with the corresponding gain.}}
\label{fig:IOU_graphs}
\end{figure}

We compare Baseline SplitFed and MuCALD SplitFed with UNet \cite{ronneberger2015UNet}, UNet3+ \cite{huang2020unet}, and SwinUNet \cite{cao_2022swin} split models. MuCALD SplitFed consistently achieves superior segmentation and privacy performance (Table~\ref{tab:BSvsMS}). Fig.~\ref{fig:IOU_graphs} shows unstable IoU N/B for Baseline SplitFed, while MuCALD SplitFed converges stably. Instability from gradient conflicts and entangled representations in Baseline SplitFed is mitigated by causal disentanglement and domain alignment. Further, comparisons are done with SoTA personalized FL approaches (FedPer \cite{arivazhagan2019federated}, FedRep \cite{collins2021exploiting}, FedBN \cite{lifedbn_2021}, FedProx \cite{li2020federated}, SCAFFOLD \cite{karimireddy2020scaffold}), and SoTA multi-task learning approaches (MOCHA \cite{zheng2020learning}, FedEM \cite{marfoq2021federated}). Table \ref{tab:qualitative_comparison} provides a qualitative comparison of the predictions. Table~\ref{tab:quantitive_comparison_combined} reports the per-client segmentation and split-point reconstruction performance. 

We further performed ablation studies with MS-UNet: (1) CRDM only, (2) DACA only, (3) disabling causal graph discovery, (4) disabling diffusion, and (5) disabling forward noising (Table~\ref{tab:ablations}). Results show that all components contribute to performance. Removing diffusion causes the largest degradation, highlighting its importance for stable and robust representations. CRDM and DACA provide complementary benefits for stability and cross-client alignment. Although disabling causal modeling increases performance in some cases, it likely reflects reliance on spurious correlations rather than robust generalization. Overall, the full model achieves the best trade-off between accuracy, stability, and privacy. Per-client ablations are available in our code repository.

\begin{table}
\small
\centering
\footnotesize
\setlength{\tabcolsep}{1.5pt}
\renewcommand{\arraystretch}{1.05}
\begin{tabular}{l|
p{0.8cm} p{0.7cm} p{0.7cm} 
p{0.7cm} p{0.7cm} p{0.75cm} 
p{0.8cm} p{0.8cm}}
\toprule
\rotatebox{00}{\textbf{Method}} &
\rotatebox{00}{\textbf{Dice}} &
\rotatebox{00}{\shortstack{\textbf{IoU} \\ \textbf{W/B}}} &
\rotatebox{00}{\shortstack{\textbf{IoU} \\ \textbf{N/B}}} &
\shortstack{\textbf{Preci}\\\textbf{\,\text{-}sion}} &
\rotatebox{00}{\textbf{Recall}} &
\rotatebox{00}{\shortstack{\textbf{F1} \\ \textbf{Score}}} &
\rotatebox{00}{\textbf{HD95}} &
\rotatebox{00}{\textbf{ASSD}} \\
%\hline
\midrule
BS-UNet &
0.394 & 0.367 & 0.045 & 0.547 & 0.435 & 0.415 & 78.911 & 31.018 \\
\textbf{MS-UNet} &
\textbf{0.816} & \textbf{0.712} & \textbf{0.579} &
\textbf{0.814} & \textbf{0.828} &
\textbf{0.816} & \textbf{37.249} & \textbf{6.203} \\ \hline
%\midrule
BS-UNet3+ &
0.680 & 0.336 & 0.593 &0.757 & 0.657 &0.680 & 45.731 & 14.615 \\
\textbf{MS-UNet3+} &
\textbf{0.909} & \textbf{0.752} & \textbf{0.842} &
\textbf{0.918} & \textbf{0.901} &
\textbf{0.909} & \textbf{20.475} & \textbf{3.339} \\ \hline
%\midrule
BS-SwinUNet &
0.384 & 0.343 & 0.044 &
0.495 & 0.400 &
0.384 & 90.5288 & 34.232 \\
\textbf{MS-SwinUNet} &
\textbf{0.733} & \textbf{0.470} & \textbf{0.626} &
\textbf{0.760} & \textbf{0.771} &
\textbf{0.733} & \textbf{44.572} & \textbf{10.546} \\
\bottomrule
\end{tabular}
\caption{\small {Average segmentation performance of the five clients across different split model architectures for Baseline SplitFed (BS) and MuCALD SplitFed (MS). Best results (MS) are shown in bold.}}
\label{tab:BSvsMS}
\end{table}
\begin{table}[!htb]
\footnotesize
\centering
\setlength{\tabcolsep}{1.5pt} 
\begin{tabular}{l|c|c|c|c|c}
\toprule
\textbf{Method} & \textbf{C\#1} & \textbf{C\#2} & \textbf{C\#3} & \textbf{C\#4} & \textbf{C\#5} \\
\midrule
\textbf{Sample} &
\predimg{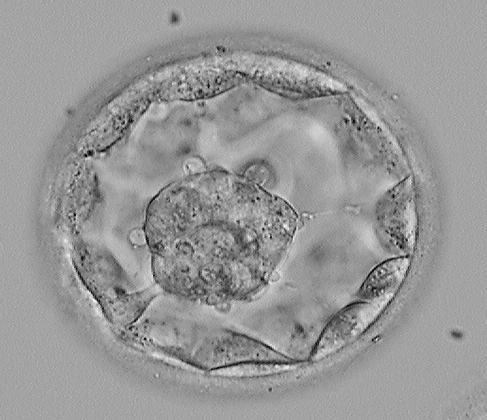} &
\predimg{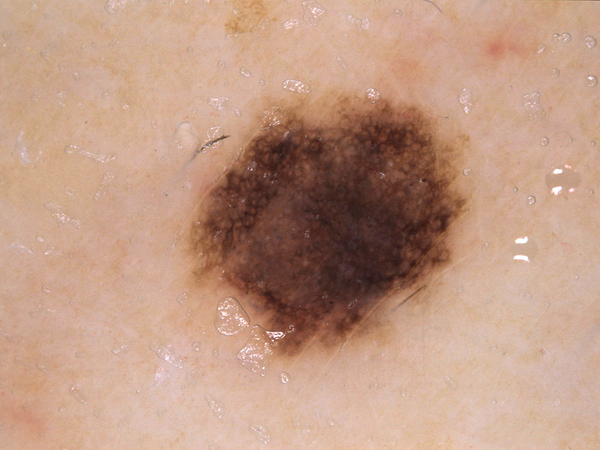} &
\predimg{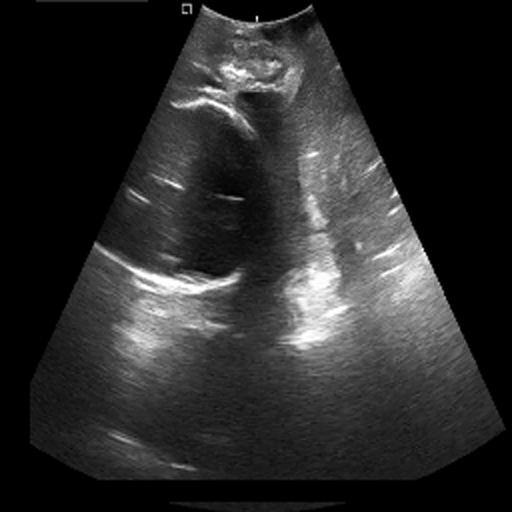} &
\predimg{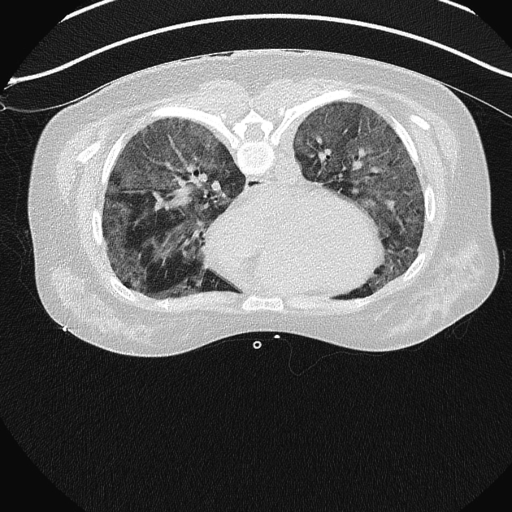} &
\predimg{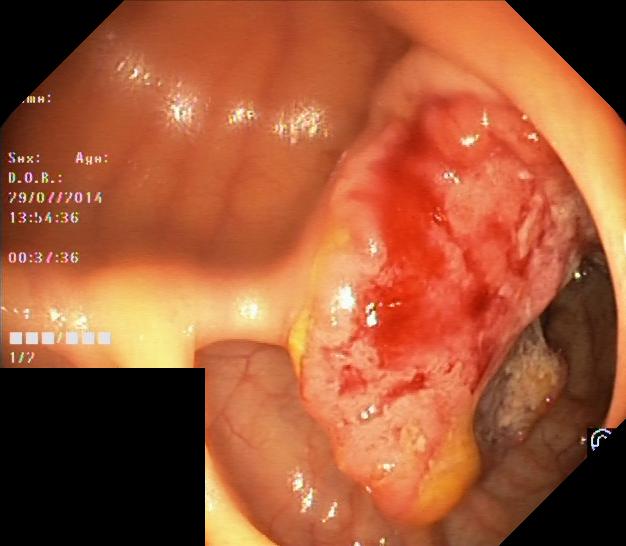} \\
\hline

\textbf{GT} &
\predimg{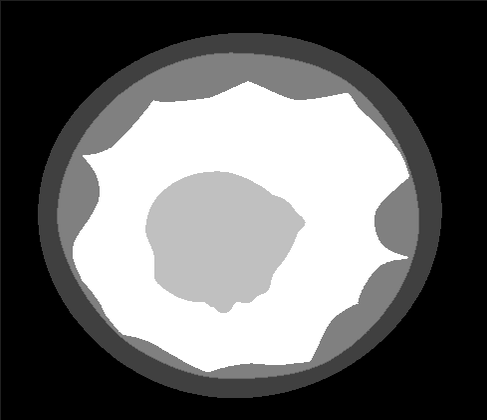} &
\predimg{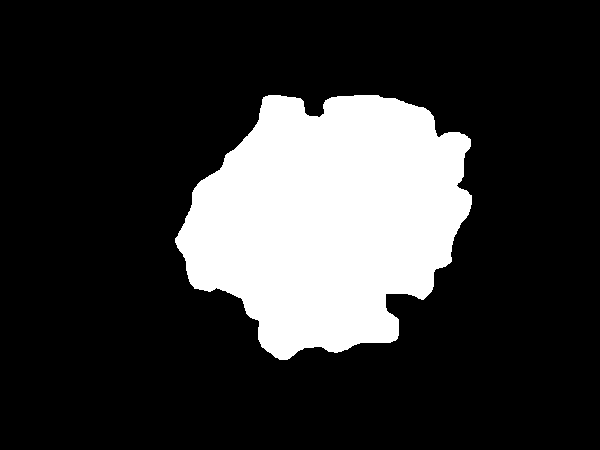} &
\predimg{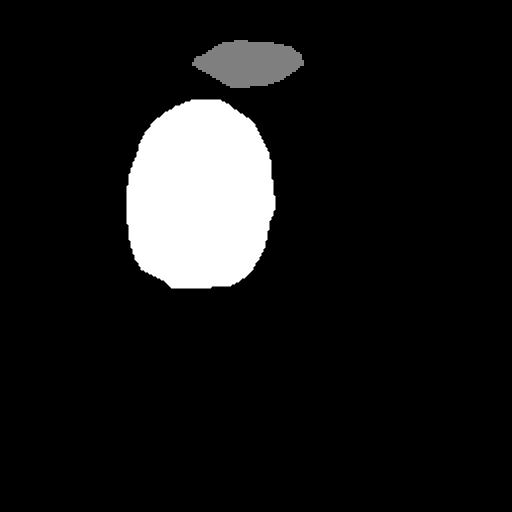} &
\predimg{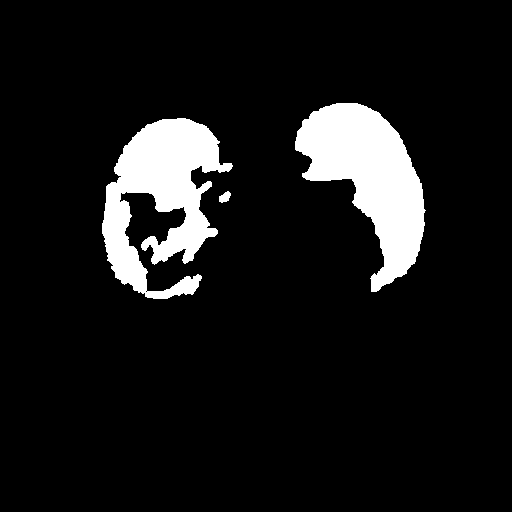} &
\predimg{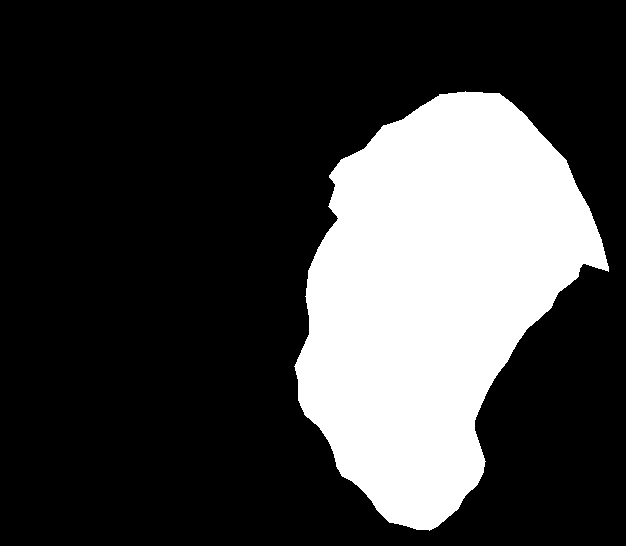} \\
\hline

\textbf{BS} &
\shortstack{\predimg{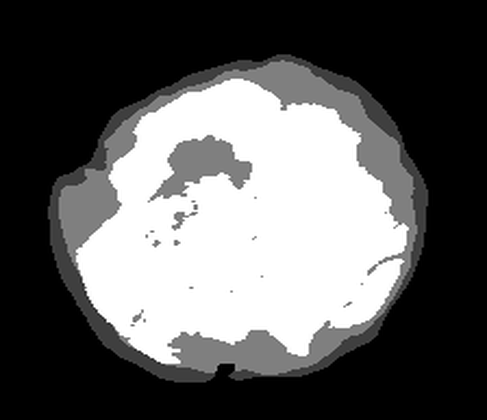}\\[-2pt]\scriptsize 0.044} &
\shortstack{\predimg{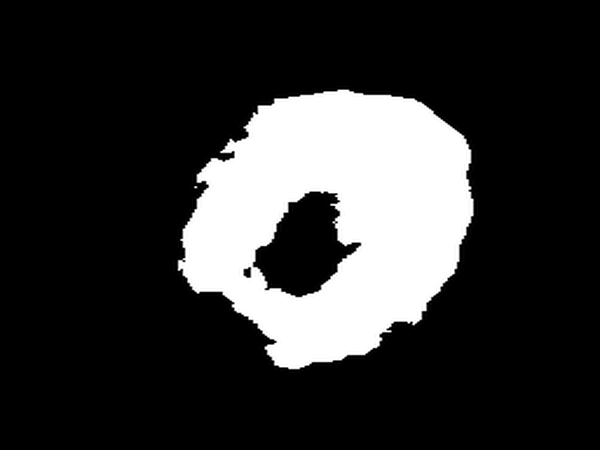}\\[-2pt]\scriptsize 0.161} &
\shortstack{\predimg{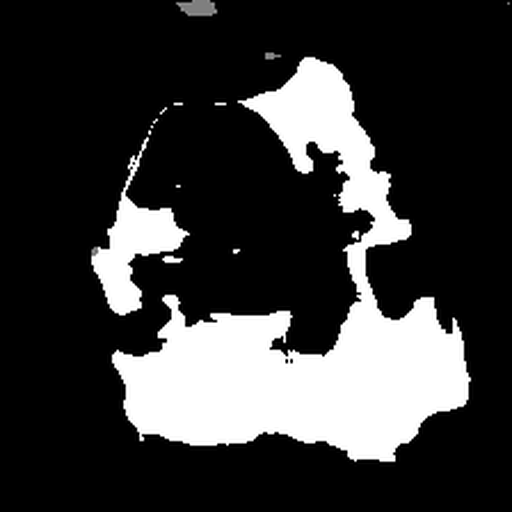}\\[-2pt]\scriptsize 0.007} &
\shortstack{\predimg{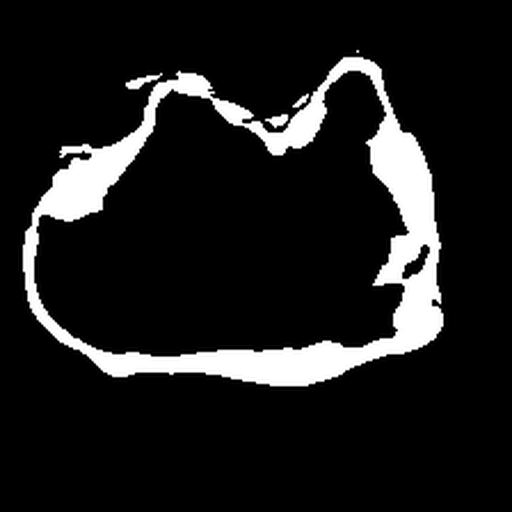}\\[-2pt]\scriptsize 0.011} &
\shortstack{\predimg{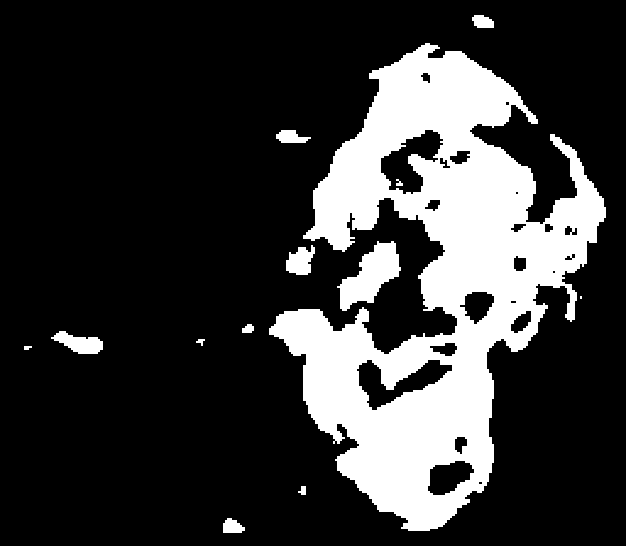}\\[-2pt]\scriptsize 0.002} \\
\hline

\textbf{FedPer} &
\shortstack{\predimg{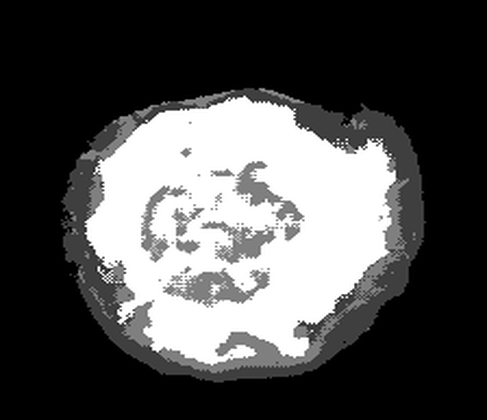}\\[-2pt]\scriptsize 0.069} &
\shortstack{\predimg{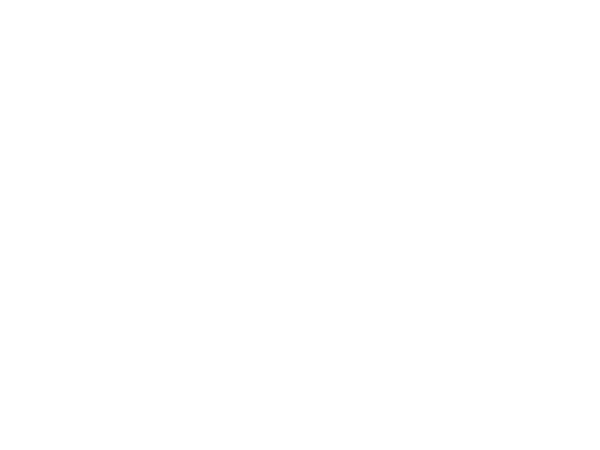}\\[-2pt]\scriptsize 0.127} &
\shortstack{\predimg{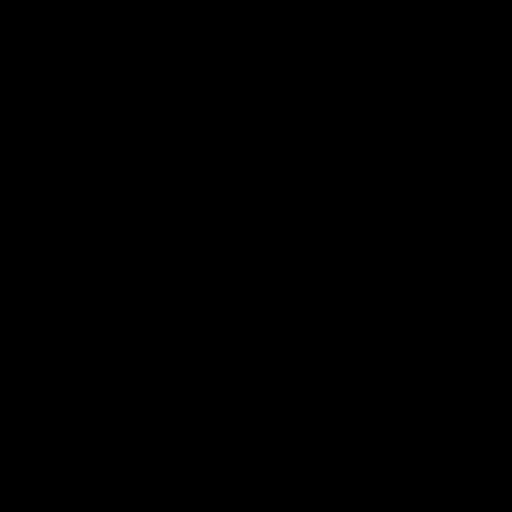}\\[-2pt]\scriptsize 0.000} &
\shortstack{\predimg{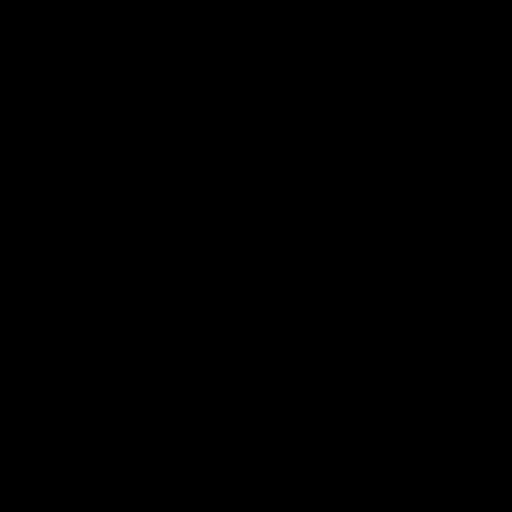}\\[-2pt]\scriptsize 0.000} &
\shortstack{\predimg{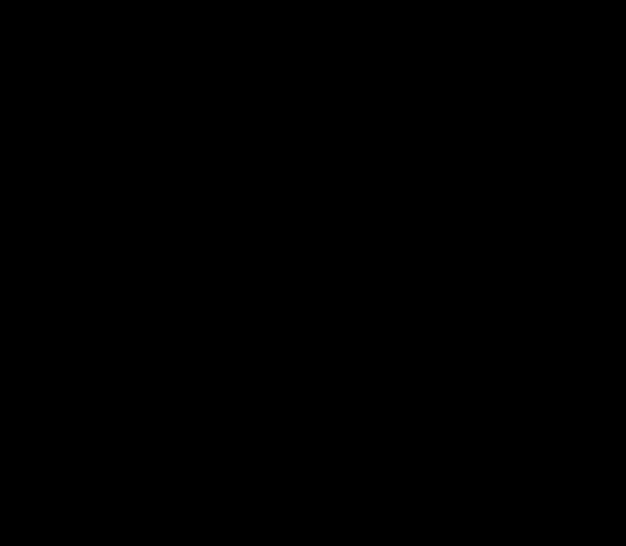}\\[-2pt]\scriptsize 0.000} \\
\hline

\textbf{FedRep} &
\shortstack{\predimg{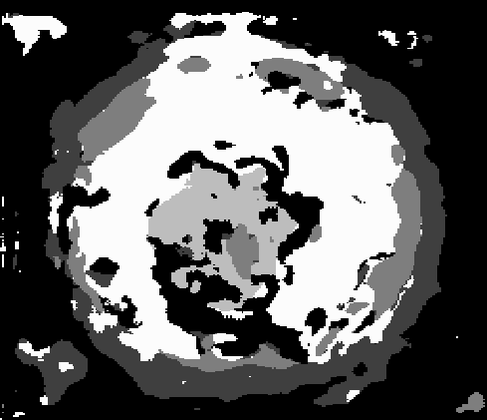}\\[-2pt]\scriptsize 0.148} &
\shortstack{\predimg{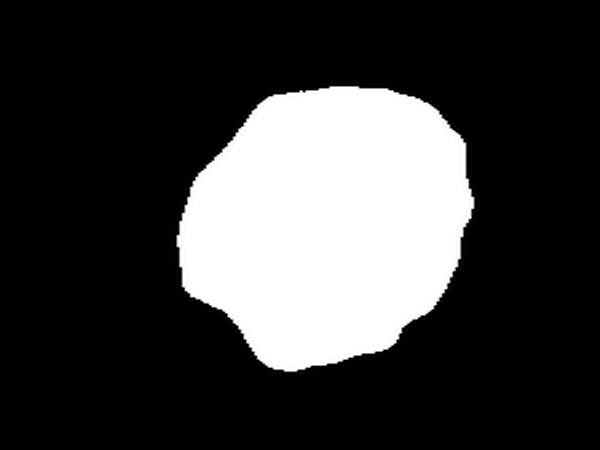}\\[-2pt]\scriptsize 0.800} &
\shortstack{\predimg{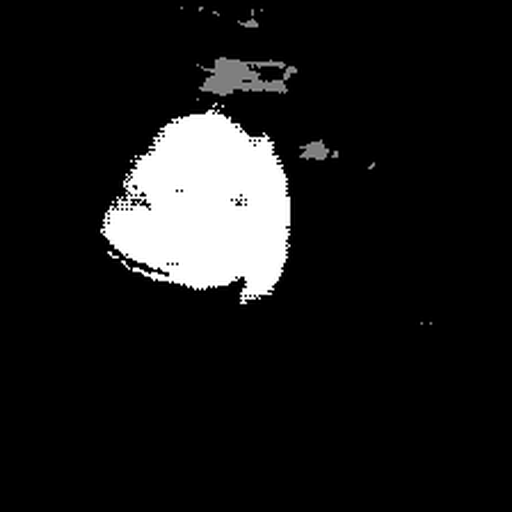}\\[-2pt]\scriptsize 0.395} &
\shortstack{\predimg{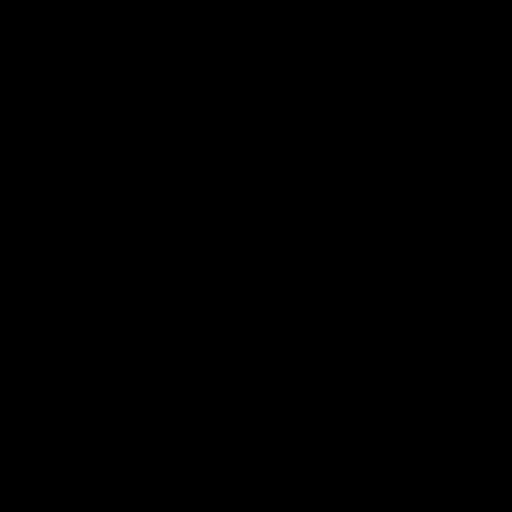}\\[-2pt]\scriptsize 0.000} &
\shortstack{\predimg{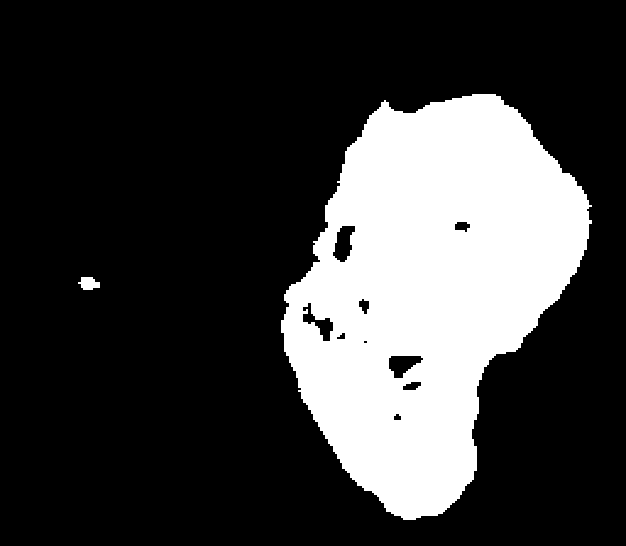}\\[-2pt]\scriptsize 0.281} \\
\hline

\textbf{FedBN} &
\shortstack{\predimg{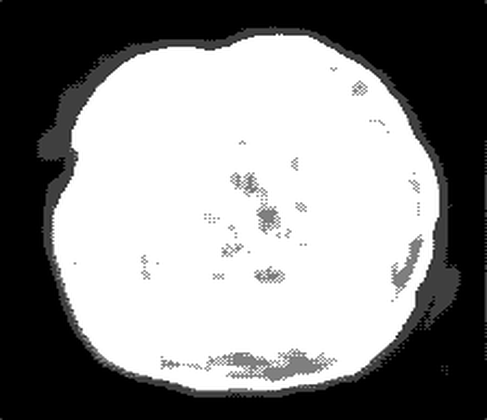}\\[-2pt]\scriptsize 0.179} &
\shortstack{\predimg{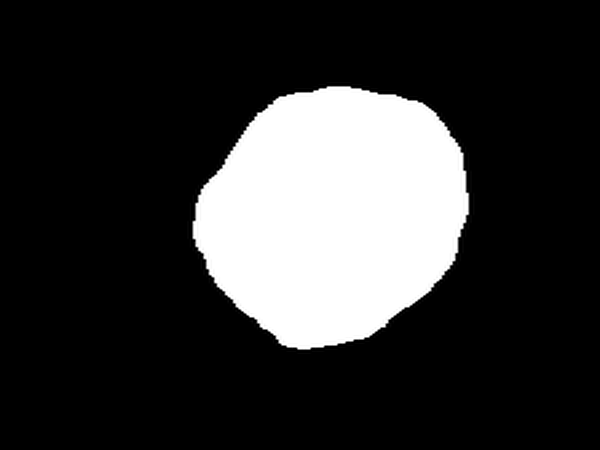}\\[-2pt]\scriptsize 0.121} &
\shortstack{\predimg{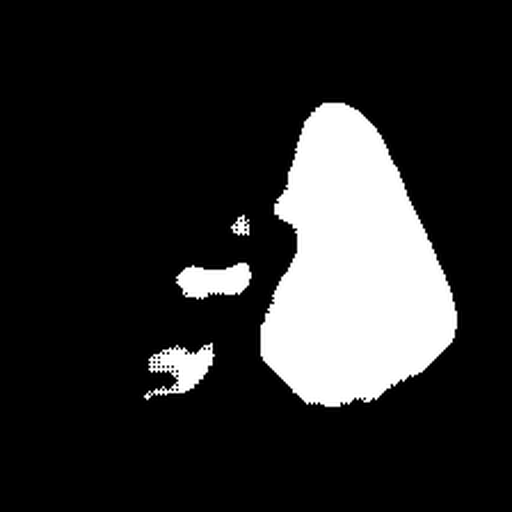}\\[-2pt]\scriptsize 0.086} &
\shortstack{\predimg{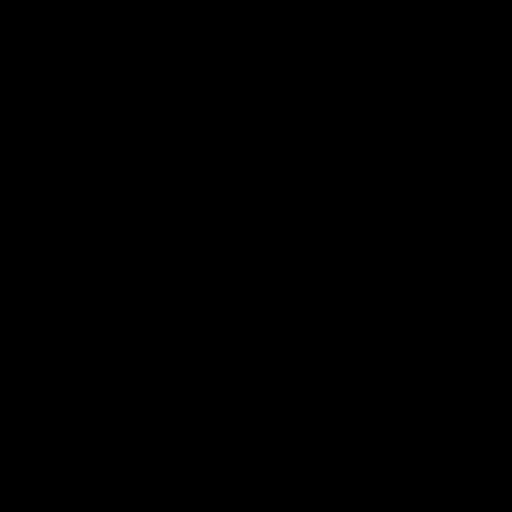}\\[-2pt]\scriptsize 0.000} &
\shortstack{\predimg{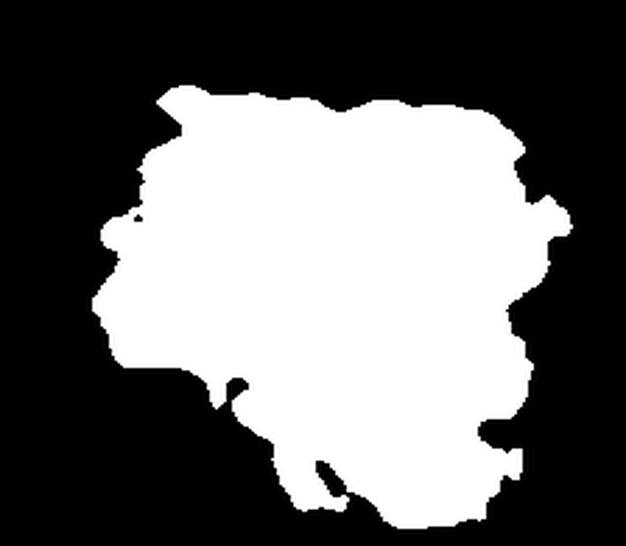}\\[-2pt]\scriptsize 0.000} \\
\hline

\textbf{FedProx} &
\shortstack{\predimg{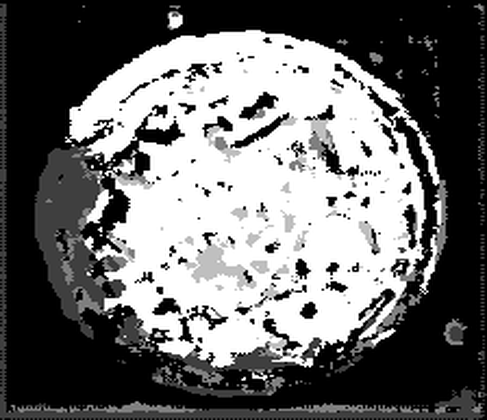}\\[-2pt]\scriptsize 0.162} &
\shortstack{\predimg{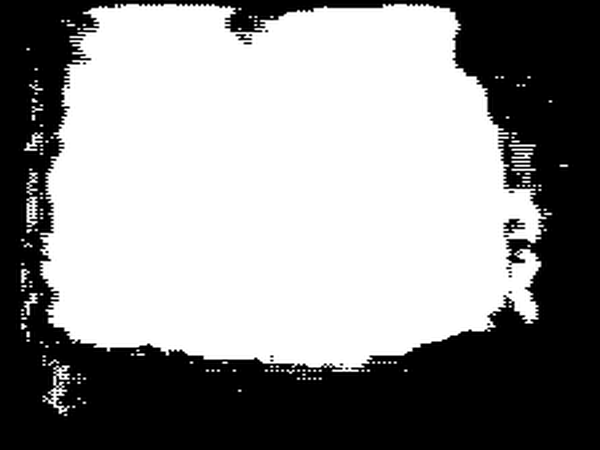}\\[-2pt]\scriptsize 0.433} &
\shortstack{\predimg{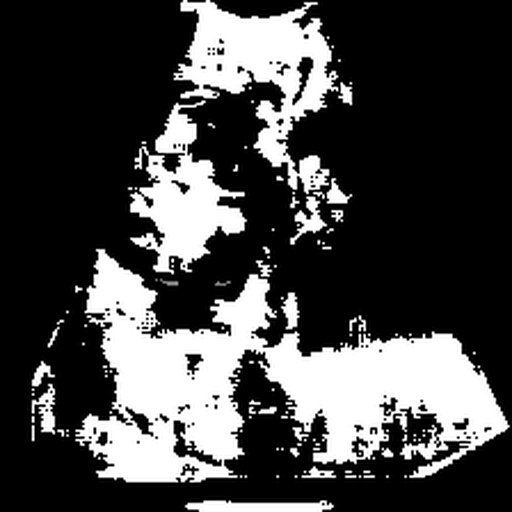}\\[-2pt]\scriptsize 0.068} &
\shortstack{\predimg{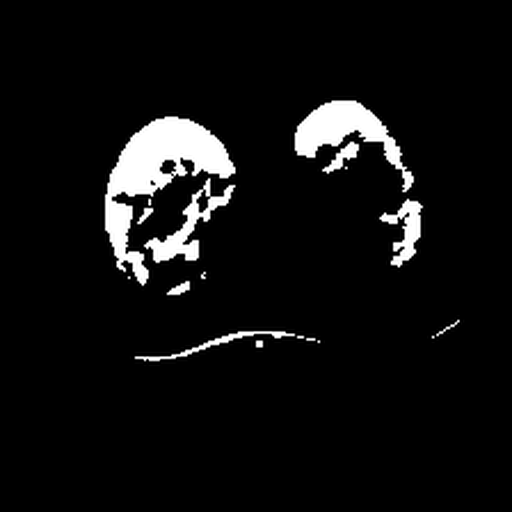}\\[-2pt]\scriptsize 0.130} &
\shortstack{\predimg{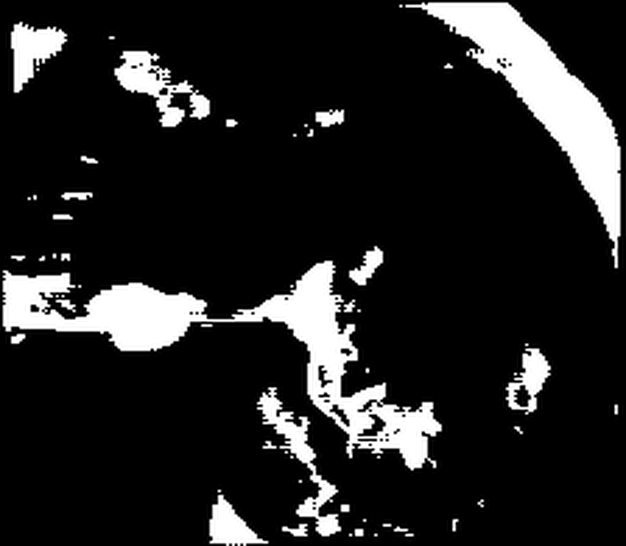}\\[-2pt]\scriptsize 0.131} \\
\hline

\textbf{SCAFFOLD} &
\shortstack{\predimg{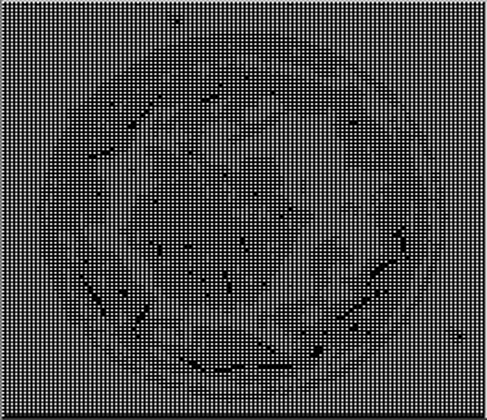}\\[-2pt]\scriptsize 0.058} &
\shortstack{\predimg{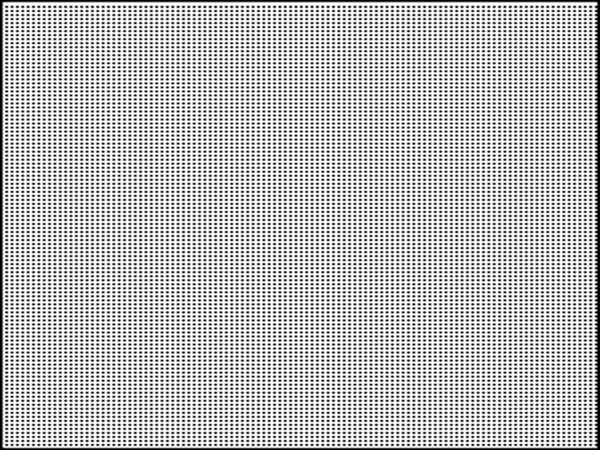}\\[-2pt]\scriptsize 0.237} &
\shortstack{\predimg{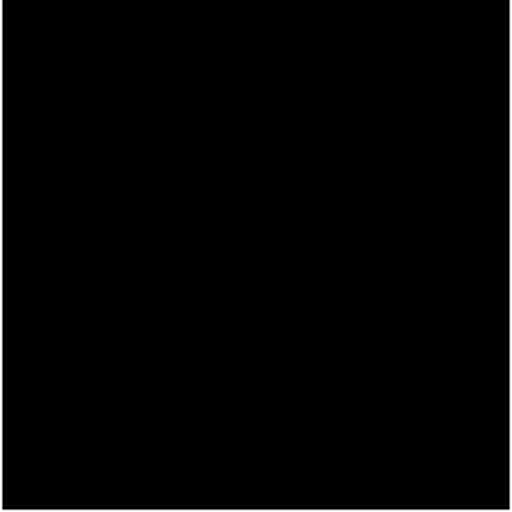}\\[-2pt]\scriptsize 0.000} &
\shortstack{\predimg{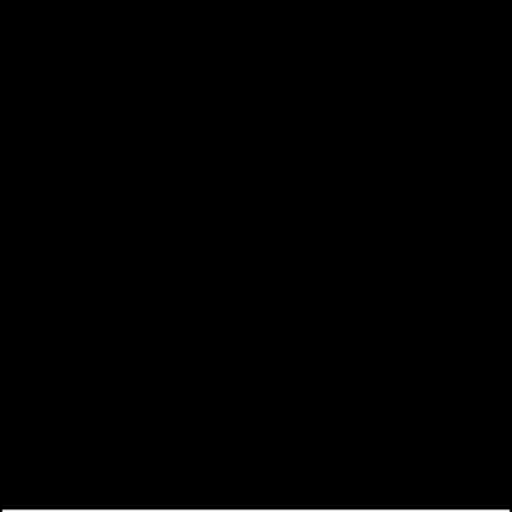}\\[-2pt]\scriptsize 0.000} &
\shortstack{\predimg{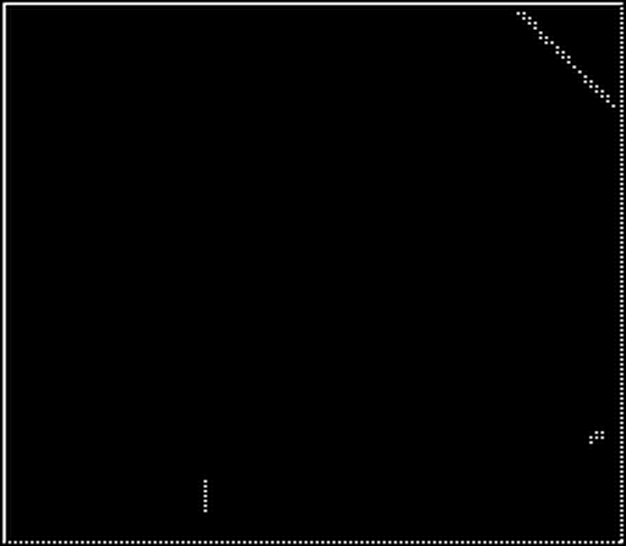}\\[-2pt]\scriptsize 0.001} \\
\hline

\textbf{MOCHA} &
\shortstack{\predimg{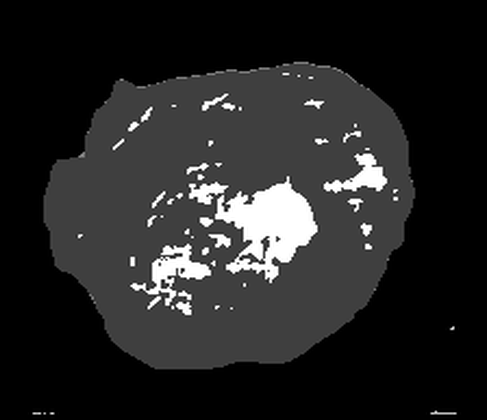}\\[-2pt]\scriptsize 0.069} &
\shortstack{\predimg{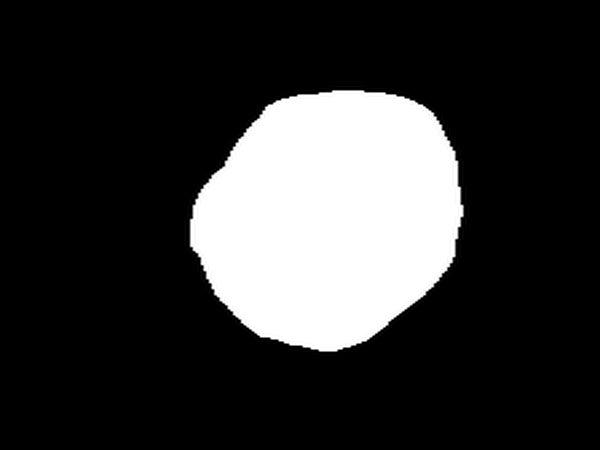}\\[-2pt]\scriptsize 0.008} &
\shortstack{\predimg{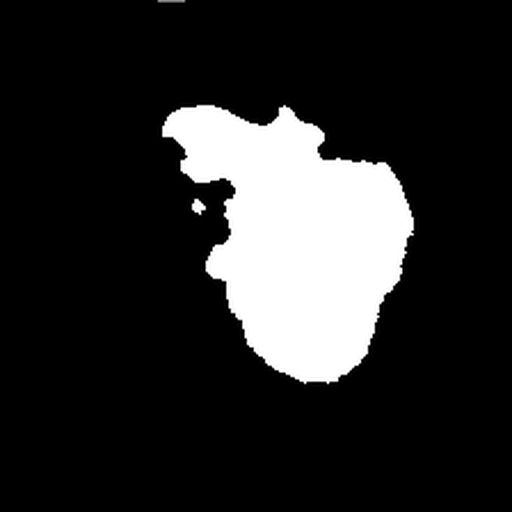}\\[-2pt]\scriptsize 0.393} &
\shortstack{\predimg{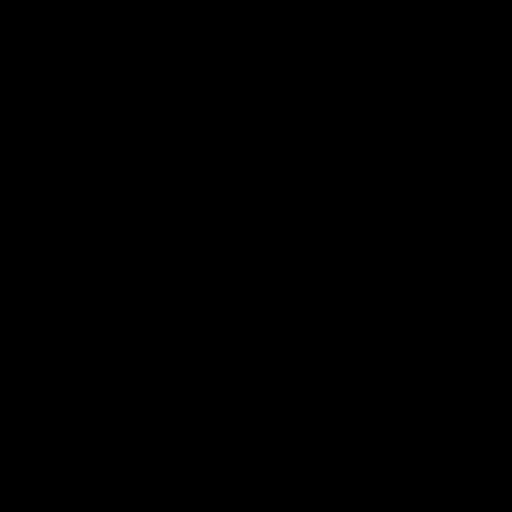}\\[-2pt]\scriptsize 0.142} &
\shortstack{\predimg{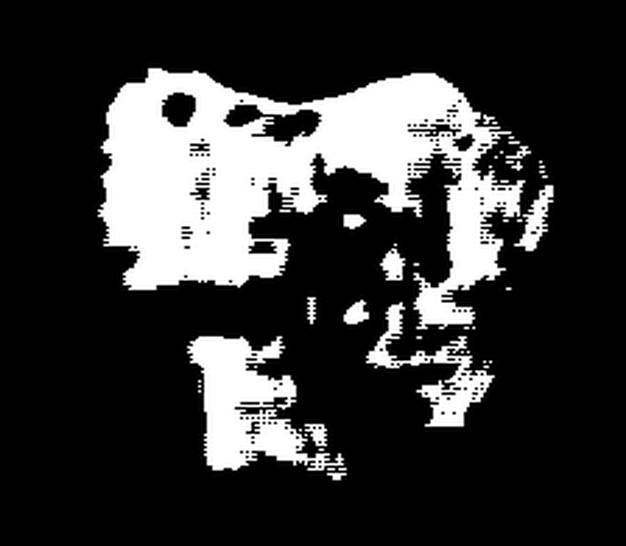}\\[-2pt]\scriptsize 0.130} \\
\hline

\textbf{FedEM} &
\shortstack{\predimg{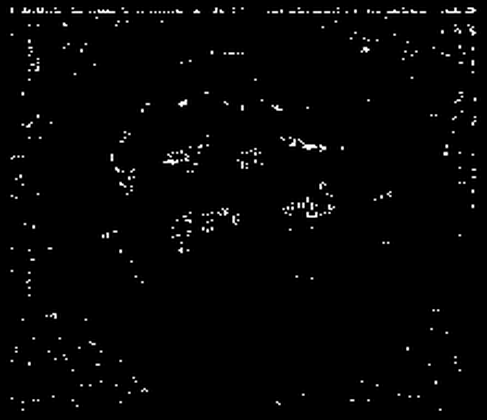}\\[-2pt]\scriptsize 0.004} &
\shortstack{\predimg{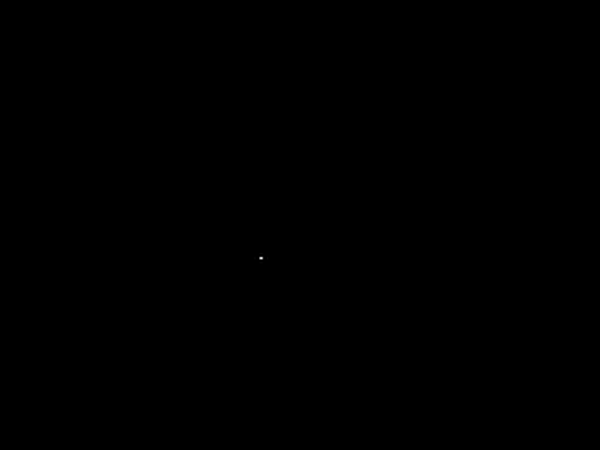}\\[-2pt]\scriptsize 0.261} &
\shortstack{\predimg{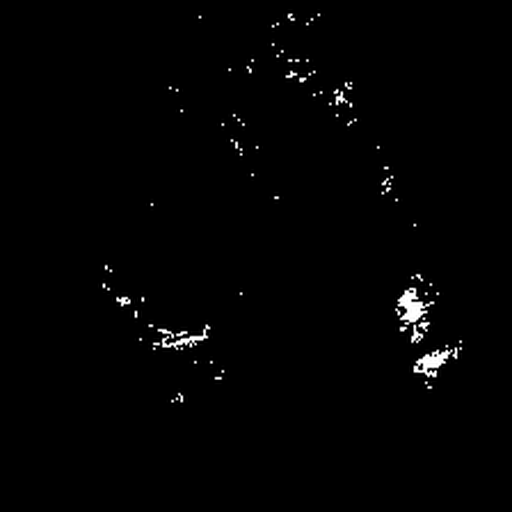}\\[-2pt]\scriptsize 0.015} &
\shortstack{\predimg{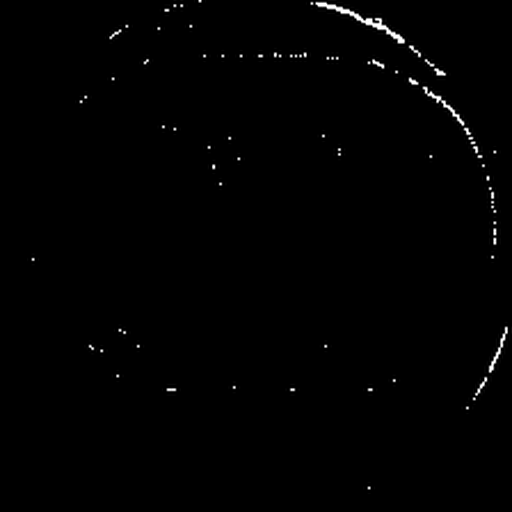}\\[-2pt]\scriptsize 0.002} &
\shortstack{\predimg{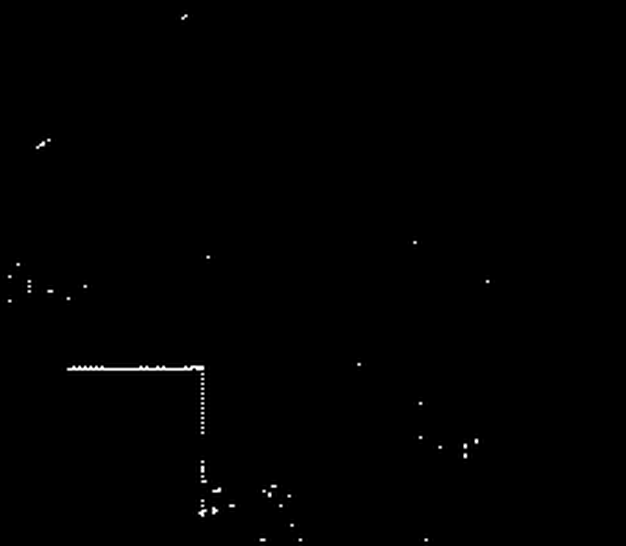}\\[-2pt]\scriptsize 0.003} \\
\hline

\textbf{MS} &
\shortstack{\predimg{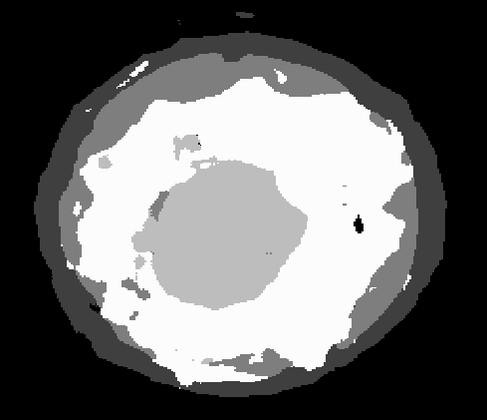}\\[-2pt]\scriptsize \textbf{0.556}} &
\shortstack{\predimg{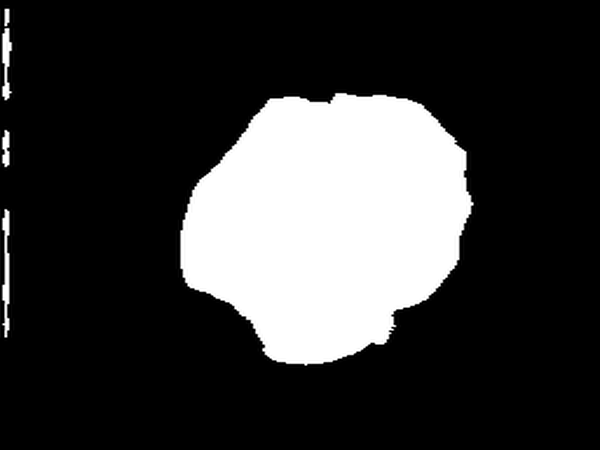}\\[-2pt]\scriptsize \textbf{0.807}} &
\shortstack{\predimg{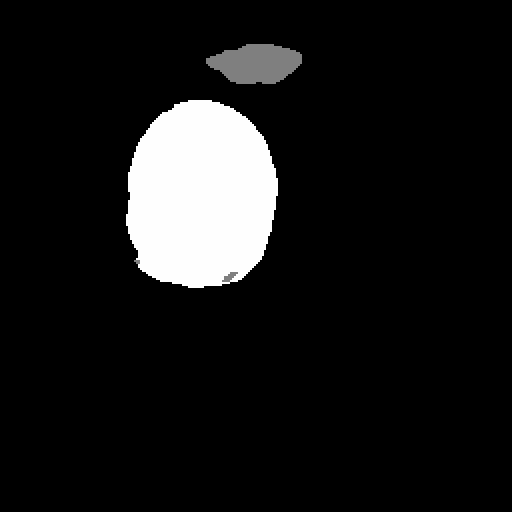}\\[-2pt]\scriptsize \textbf{0.620}} &
\shortstack{\predimg{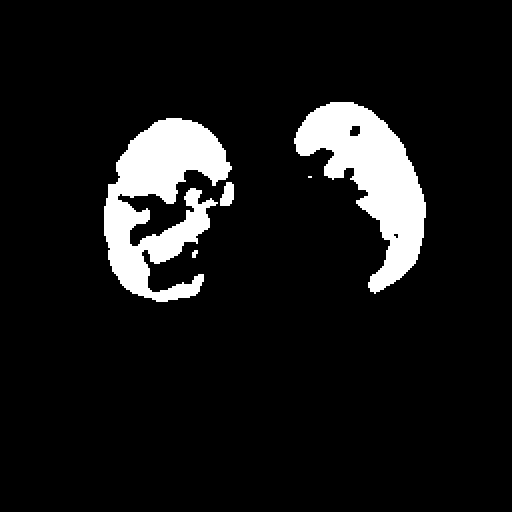}\\[-2pt]\scriptsize \textbf{0.442}} &
\shortstack{\predimg{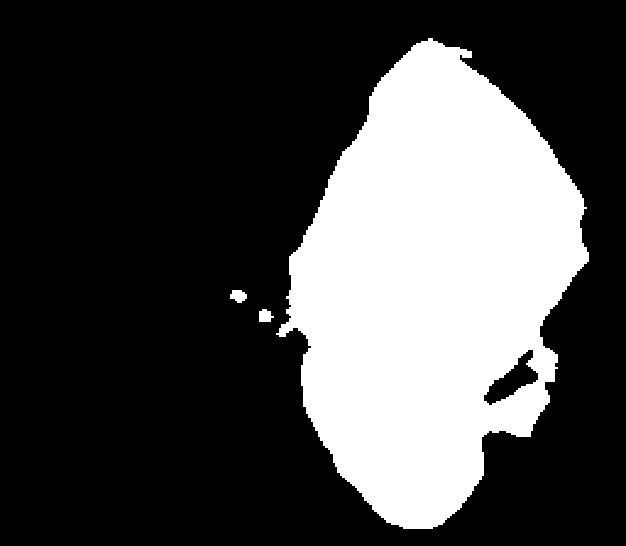}\\[-2pt]\scriptsize \textbf{0.468}} \\
\bottomrule
\end{tabular}
\caption{\small{Qualitative segmentation comparison across five clients using one representative test sample per client, with ground truth and client-level test-set IoU N/B.}}
\label{tab:qualitative_comparison}
\end{table}

\section{Conclusion \& Future Works}
\label{sec:conclusion}
This work introduced MuCALD-SplitFed, a causal–latent diffusion framework addressing stability, privacy, and cross-task generalization in multi-task SplitFed. By combining causal representation learning, diffusion-based obfuscation, and domain-adversarial alignment, MuCALD-SplitFed consistently improves segmentation performance across five medical imaging datasets, outperforming Baseline SplitFed and SoTA personalized/ multi-task FL methods. Reconstruction metrics show feature maps are far more obfuscated, making reconstruction and membership-inference attacks effectively infeasible. This work presents an initial step toward comprehensive privacy-preserving multi-task SplitFed learning. Future work will evaluate stronger adversarial models, assess scalability to larger and more diverse client populations, and extend to broader causal settings, additional modalities, and adaptive privacy under increased heterogeneity. 

\begin{table*}[!t]
\centering
%\footnotesize
\scriptsize
%\tiny
\setlength{\tabcolsep}{2pt} 
\renewcommand{\arraystretch}{1.05}
\begin{tabular}{l|l|cccccccc|ccc|ccc}
\toprule
\textbf{Method} & {\textbf{Client\#}} & \multicolumn{8}{c}{\textbf{Segmentation metrics}} & \multicolumn{6}{|c}{\textbf{Reconstruction-quality metrics}} \\
\cmidrule(lr){3-10} \cmidrule(lr){11-16}
& & {\textbf{Dice}} &
{\shortstack{\textbf{IoU W/B}}} &
{\shortstack{\textbf{IoU N/B}}} &
{\textbf{Precision}} &
{\textbf{Recall}} &
{\shortstack{\textbf{F1 Score}}} &
{\textbf{HD95}} &
{\textbf{ASSD}}
& \shortstack{\textbf{MSE(S1)}}
& \shortstack{\textbf{PSNR(S1)}}
& \shortstack{\textbf{SSIM(S1)}}
& \shortstack{\textbf{MSE(S2)}}
& \shortstack{\textbf{PSNR(S2)}}
& \shortstack{\textbf{SSIM(S2)}} \\
\midrule
\multirow{6}{*}{\textbf{BS}}
& \textbf{C\#1} & 0.215 & 0.154 & 0.044 & 0.264 & 0.257 & 0.215 & 69.301 & 24.389 & 0.323 & 5.092 & 0.065 & 0.255 & 0.131 & 0.125 \\
& \textbf{C\#2} & 0.469 & 0.469 & 0.161 & 0.876 & 0.580 & 0.575 & 71.004 & 22.847 & 0.468 & 3.411 & 0.005 & 0.468 & 3.416 & 0.005 \\
& \textbf{C\#3} & 0.316 & 0.289 & 0.007 & 0.634 & 0.338 & 0.316 & 110.546 & 41.626 & 0.037 & 14.727 & 0.389 & 0.037 & 14.782 & 0.394 \\
& \textbf{C\#4} & 0.503 & 0.488 & 0.011 & 0.502 & 0.503 & 0.503 & 75.422 & 33.197 & 0.155 & 9.271 & 0.145 & 0.154 & 9.286 & 0.145 \\
& \textbf{C\#5} & 0.466 & 0.434 & 0.002 & 0.459 & 0.498 & 0.466 & 68.281 & 33.029 & 0.218 & 6.716 & 0.115 & 0.175 & 7.712 & 0.213 \\
\cline{2-16}
& \textbf{Avg}  & 0.394 & 0.367 & 0.045 & 0.547 & 0.435 & 0.415 & 78.911 & 31.018 & 0.240 & 7.843 & 0.144 & 0.218 & 7.059 & 0.177 \\
\hline
%\midrule
\multirow{6}{*}{\textbf{FedPer}}
& \textbf{C\#1} & 0.086 & 0.055 & 0.069 & 0.255 & 0.200 & 0.086 & 212.194 & 73.481 & 0.143 & 9.035 & 0.420 & 0.116 & 10.055 & 0.401 \\
& \textbf{C\#2} & 0.203 & 0.254 & 0.127 & 0.127 & 0.500 & 0.203 & 196.640 & 61.106 & 0.173 & 8.114 & 0.475 & 0.106 & 10.496 & 0.491 \\
& \textbf{C\#3} & 0.307 & 0.284 & 0.000 & 0.284 & 0.333 & 0.307 & 125.446 & 46.152 & 0.036 & 14.826 & 0.389 & 0.036 & 14.826 & 0.389 \\
& \textbf{C\#4} & 0.496 & 0.492 & 0.000 & 0.492 & 0.500 & 0.496 & 95.260 & 40.435 & 0.159 & 9.234 & 0.142 & 0.159 & 9.234 & 0.142 \\
& \textbf{C\#5} & 0.465 & 0.435 & 0.000 & 0.435 & 0.500 & 0.465 & 113.236 & 45.931 & 0.218 & 6.739 & 0.115 & 0.218 & 6.739 & 0.115 \\
\cline{2-16}
& \textbf{Avg} & 0.311 & 0.304 & 0.039 & 0.319 & 0.407 & 0.311 & 148.555 & 53.421 & 0.145 & 9.610 & 0.308 & 0.127 & 10.270 & 0.307 \\
\hline
%\midrule
\multirow{6}{*}{\textbf{FedRep}}
& \textbf{C\#1} & 0.347 & 0.244 & 0.148 & 0.483 & 0.366 & 0.347 & 46.694 & 11.824 & 0.338 & 4.985 & 0.063 & 0.245 & 6.386 & 0.168 \\
& \textbf{C\#2} & 0.925 & 0.863 & 0.800 & 0.922 & 0.928 & 0.925 & 14.498 & 2.438 & 0.474 & 3.316 & 0.053 & 0.467 & 3.396 & 0.056 \\
& \textbf{C\#3} & 0.672 & 0.571 & 0.395 & 0.754 & 0.622 & 0.672 & 38.063 & 7.200 & 0.103 & 10.138 & 0.389 & 0.096 & 10.472 & 0.392 \\
& \textbf{C\#4} & 0.496 & 0.492 & 0.000 & 0.507 & 0.500 & 0.496 & 88.869 & 39.609 & 0.159 & 9.234 & 0.142 & 0.157 & 9.257 & 0.177 \\
& \textbf{C\#5} & 0.689 & 0.583 & 0.281 & 0.773 & 0.653 & 0.689 & 41.971 & 13.546 & 0.207 & 6.977 & 0.133 & 0.179 & 7.618 & 0.141 \\
\cline{2-16}
& \textbf{Avg} & 0.626 & 0.551 & 0.325 & 0.688 & 0.614 & 0.626 & 46.019 & 14.923 & 0.256 & 6.744 & 0.156 & 0.228 & 7.226 & 0.187 \\
\hline
%\midrule
\multirow{6}{*}{\textbf{FedBN}}
& \textbf{C\#1} & 0.374 & 0.277 & 0.179 & 0.377 & 0.401 & 0.374 & 45.935 & 16.876 & 0.290 & 5.401 & 0.096 & 0.215 & 6.758 & 0.110 \\
& \textbf{C\#2} & 0.542 & 0.445 & 0.121 & 0.876 & 0.560 & 0.542 & 82.266 & 27.963 & 0.471 & 3.395 & 0.003 & 0.469 & 3.412 & 0.004 \\
& \textbf{C\#3} & 0.400 & 0.333 & 0.086 & 0.417 & 0.393 & 0.400 & 92.351 & 30.747 & 0.062 & 12.688 & 0.397 & 0.058 & 12.975 & 0.398 \\
& \textbf{C\#4} & 0.496 & 0.492 & 0.000 & 0.492 & 0.500 & 0.496 & 95.260 & 40.435 & 0.159 & 9.234 & 0.142 & 0.158 & 9.241 & 0.144 \\
& \textbf{C\#5} & 0.466 & 0.435 & 0.000 & 0.902 & 0.500 & 0.466 & 111.620 & 45.246 & 0.218 & 6.739 & 0.115 & 0.143 & 8.578 & 0.214 \\
\cline{2-16}
& \textbf{Avg}  & 0.456 & 0.396 & 0.077 & 0.613 & 0.471 & 0.456 & 85.486 & 32.253 & 0.240 & 8.491 & 0.151 & 0.209 & 8.592 & 0.174 \\
\hline
%\midrule
\multirow{6}{*}{\textbf{FedProx}}
& \textbf{C\#1} & 0.317 & 0.211 & 0.162 & 0.327 & 0.323 & 0.317 & 53.136 & 11.912 & 0.283 & 5.704 & 0.071 & 0.217 & 6.794 & 0.034 \\
& \textbf{C\#2} & 0.690 & 0.533 & 0.433 & 0.699 & 0.762 & 0.690 & 45.377 & 9.486 & 0.391 & 4.324 & 0.158 & 0.368 & 4.648 & 0.143 \\
& \textbf{C\#3} & 0.367 & 0.297 & 0.068 & 0.383 & 0.377 & 0.367 & 69.066 & 18.966 & 0.137 & 8.827 & 0.400 & 0.113 & 9.665 & 0.404 \\
& \textbf{C\#4} & 0.609 & 0.553 & 0.130 & 0.608 & 0.610 & 0.609 & 38.697 & 12.701 & 0.164 & 9.125 & 0.141 & 0.159 & 9.296 & 0.148 \\
& \textbf{C\#5} & 0.562 & 0.469 & 0.131 & 0.566 & 0.559 & 0.562 & 60.848 & 16.104 & 0.189 & 7.355 & 0.133 & 0.141 & 8.643 & 0.170 \\
\cline{2-16}
& \textbf{Avg}  & 0.509 & 0.413 & 0.185 & 0.517 & 0.526 & 0.509 & 53.425 & 13.834 & 0.232 & 7.009 & 0.180 & 0.199 & 7.405 & 0.180 \\
\hline
%\midrule
\multirow{6}{*}{\textbf{SCAFFOLD}}
& \textbf{C\#1} & 0.168 & 0.100 & 0.058 & 0.160 & 0.188 & 0.168 & 104.390 & 41.987 & 0.330 & 4.893 & 0.009 & 0.044 & 14.302 & 0.096 \\
& \textbf{C\#2} & 0.388 & 0.240 & 0.237 & 0.507 & 0.507 & 0.388 & 9.823 & 52.924 & 0.249 & 6.074 & 0.003 & 0.200 & 7.020 & 0.003 \\
& \textbf{C\#3} & 0.304 & 0.280 & 0.000 & 0.284 & 0.328 & 0.304 & 105.006 & 53.213 & 0.050 & 13.265 & 0.377 & 0.047 & 13.494 & 0.353 \\
& \textbf{C\#4} & 0.495 & 0.490 & 0.000 & 0.492 & 0.498 & 0.495 & 77.754 & 53.558 & 0.163 & 9.034 & 0.139 & 0.162 & 9.059 & 0.141 \\
& \textbf{C\#5} & 0.462 & 0.429 & 0.001 & 0.440 & 0.493 & 0.462 & 72.343 & 34.309 & 0.226 & 6.567 & 0.100 & 0.126 & 9.096 & 0.023 \\
\cline{2-16}
& \textbf{Avg}  & 0.363 & 0.308 & 0.059 & 0.377 & 0.403 & 0.363 & 73.863 & 47.198 & 0.203 & 7.754 & 0.126 & 0.115 & 10.594 & 0.122 \\
\hline
%\midrule
\multirow{6}{*}{\textbf{MOCHA}}
& \textbf{C\#1} & 0.217 & 0.164 &0.037 & 0.350 & 0.249  & 0.217  & 66.205 & 26.965 & 0.334 & 4.995 & 0.069 & 0.171 & 7.936 & 0.230 \\
& \textbf{C\#2} & 0.620  &0.503  &0.218  &0.875  & 0.608  & 0.620  & 61.653 & 18.819 & 0.473 & 3.364 & 0.008 & 0.472 & 3.376 & 0.008 \\
& \textbf{C\#3} &  0.420 &0.352   &0.102  &0.576  &0.405  &  0.420 & 85.198 & 26.946 & 0.060 & 12.985 & 0.393 & 0.059 & 13.079 & 0.366 \\
& \textbf{C\#4} & 0.496 & 0.492  &0.000  &0.491 &0.500   &0.496   & 95.260 & 40.435 & 0.159 & 9.234 & 0.142 & 0.159 & 9.235 & 0.142 \\
& \textbf{C\#5}& 0.602  &0.503   &0.172  & 0.618 & 0.592  &  0.602 & 43.753 & 12.004 & 0.219 & 6.708 & 0.130 & 0.154 & 8.224 & 0.141 \\
\cline{2-16}
& \textbf{Avg} & 0.471  &0.403  &0.106 & 0.582& 0.471 & 0.471  &70.414 & 25.034 & 0.248 & 7.257 & 0.148 & 0.159 & 8.770 & 0.178 \\
\hline
%\midrule
\multirow{6}{*}{\textbf{FedEM}}
& \textbf{C\#1} & 0.178 & 0.074 & 0.004 & 0.160 & 0.178 & 0.152 & 107.140 & 41.681 & 0.434 & 3.821 & 0.001 & 0.260 & 6.462 & 0.096 \\
& \textbf{C\#2} & 0.473 & 0.517 & 0.261 & 0.507 & 0.518 & 0.368 & 9.682 & 53.124 & 0.471 & 3.375 & 0.009 & 0.329 & 5.312 & 0.057 \\
& \textbf{C\#3} & 0.150 & 0.291 & 0.015 & 0.284 & 0.314 & 0.314 & 103.006 & 54.213 & 0.032 & 15.290 & 0.393 & 0.014 & 18.840 & 0.303 \\
& \textbf{C\#4} & 0.495 & 0.486 & 0.002 & 0.492 & 0.499 & 0.494 & 76.754 & 55.558 & 0.169 & 8.679 & 0.127 & 0.112 & 10.174 & 0.063 \\
& \textbf{C\#5} & 0.513 & 0.435 & 0.003 & 0.440 & 0.495 & 0.482 & 70.344 & 31.309 & 0.219 & 6.708 & 0.112 & 0.163 & 7.993 & 0.098 \\
\cline{2-16}
& \textbf{Avg}  & 0.362 & 0.361 & 0.057 & 0.377 & 0.401 & 0.362 & 73.385 & 47.177 & 0.264 & 7.774 & 0.187 & 0.156 & 9.297 & 0.124 \\
\hline
%\midrule
\multirow{6}{*}{\textbf{MS}}
& \textbf{C\#1} & \textbf{0.728} & \textbf{0.574} & \textbf{0.556} & \textbf{0.739} & \textbf{0.721} & \textbf{0.728} & \textbf{45.730} & \textbf{6.717} & \textbf{0.430} & \textbf{35.264} & \textbf{0.790} & \textbf{0.110} & \textbf{35.544} & \textbf{0.748} \\
& \textbf{C\#2} & \textbf{0.929} & \textbf{0.869} & \textbf{0.807} & \textbf{0.930} & \textbf{0.927} & \textbf{0.929} & \textbf{26.008} & \textbf{3.835} & \textbf{0.238} & \textbf{32.753} & \textbf{0.467} & \textbf{0.071} & \textbf{40.285} & \textbf{0.764} \\
& \textbf{C\#3} & \textbf{0.831} & \textbf{0.726} & \textbf{0.620} & \textbf{0.867} & \textbf{0.816} & \textbf{0.831} & \textbf{47.909} & \textbf{8.815} & \textbf{0.314} & \textbf{38.560} & \textbf{0.770} & \textbf{0.082} & \textbf{40.298} & \textbf{0.816} \\
& \textbf{C\#4} & \textbf{0.802} & \textbf{0.713} & \textbf{0.442} & \textbf{0.755} & \textbf{0.874} & \textbf{0.802} & \textbf{28.834} & \textbf{5.003} & \textbf{0.422} & \textbf{36.594} & \textbf{0.663} & \textbf{0.119} & \textbf{35.709} & \textbf{0.742} \\
& \textbf{C\#5} & \textbf{0.790} & \textbf{0.680} & \textbf{0.468} & \textbf{0.780} & \textbf{0.802} & \textbf{0.790} & \textbf{37.762} & \textbf{6.646} & \textbf{0.266} & \textbf{34.302} & \textbf{0.561} & \textbf{0.092} & \textbf{38.277} & \textbf{0.756} \\
\cline{2-16}
& \textbf{Avg}  & \textbf{0.816} & \textbf{0.712} & \textbf{0.579} & \textbf{0.814} & \textbf{0.828} & \textbf{0.816} & \textbf{37.249} & \textbf{6.203} & \textbf{0.334} & \textbf{35.695} & \textbf{0.650} & \textbf{0.095} & \textbf{38.823} & \textbf{0.765} \\
\bottomrule
\end{tabular}
\caption{\small{Segmentation and reconstruction-quality metrics over the 5 clients for all methods. BS: Baseline SplitFed, MS: MuCALD SplitFed, S1: Split-1, S2: Split-2, Best values (MS) are in bold.}}
\label{tab:quantitive_comparison_combined}
\end{table*}
\vspace{-2cm}
\begin{table*}[!b]
\centering
%\footnotesize
\scriptsize
\setlength{\tabcolsep}{2pt}
\renewcommand{\arraystretch}{1.05}
\begin{tabular}{l|cccccccc|ccc|ccc}
\toprule
\textbf{Ablation study} & \multicolumn{8}{c|}{\textbf{Segmentation metrics}} & \multicolumn{6}{c}{\textbf{Reconstruction-quality metrics}} \\
\cmidrule(lr){2-9} \cmidrule(lr){10-15}
& \textbf{Dice}
& \shortstack{\textbf{IoU W/B}}
& \shortstack{\textbf{IoU N/B}}
& \textbf{Precision}
& \textbf{Recall}
& \shortstack{\textbf{F1 Score}}
& \textbf{HD95}
& \textbf{ASSD}
& \shortstack{\textbf{MSE(S1)}}
& \shortstack{\textbf{PSNR(S1)}}
& \shortstack{\textbf{SSIM(S1)}}
& \shortstack{\textbf{MSE(S2)}}
& \shortstack{\textbf{PSNR(S2)}}
& \shortstack{\textbf{SSIM(S2)}} \\
\midrule
\textbf{CRDM only} & 0.814 &0.720  &0.579  & 0.858& 0.801&0.814  &27.957  &5.551  & 1.175 & 53.351 &0.922  &0.116  & 36.608 & 0.758   \\
\textbf{DACA only} &0.832  &0.734  &0.607  &0.850  & 0.837 & 0.832 &30.489  & 5.272 & --- & --- &---  & --- &---  & --- \\
\textbf{Causality disabled} &0.843  &0.747  &0.627  &0.857  &0.844  & 0.843 &31.544  &5.251  & 1.151 & 53.093 & 0.940 & 0.132 & 36.120 &0.771  \\
\textbf{Diffusion disabled} & 0.800 &0.689  &0.552  &0.811  &0.817  & 0.800&44.269 & 7.264 & ---  & ---  &  --- &---   & ---  & ---  \\
\textbf{Forward noise disabled} &0.785  & 0.683 &  0.533& 0.808 &0.792  &0.785  &46.239  & 8.790 &---   &---   &---   & ---  &---   &---   \\
\bottomrule
\end{tabular}
\caption{\small Average segmentation performance for ablation experiments with MS-UNet. S1: Split-1, S2: Split-2. Reconstruction metrics are not applicable in some cases due to architectural modifications in the ablated models.}
\label{tab:ablations}
\end{table*}
%\FloatBarrier
%\newpage
\clearpage
\small{
\bibliographystyle{IEEEbib}
\bibliography{refs}
}

\end{document}